# *HazeSpikeMamba: Coupling Spiking-Inspired and State-Space Features for Self-Supervised Real-World Dehazing*

Haoran Liu[1,2], Li Xu[1], Mingzhe Liu[1,2,*], Huibin Li[2], Peng Li[2], Guibin Zan[3], Simon X. Yang[4]

1 School of Artificial Intelligence and Electronic Engineering, Sichuan Technology and Business University, Chengdu 611745, China

2 Chengdu University of Technology, College of Nuclear Technology and Automation Engineering, Chengdu 610059, China

3 National Synchrotron Radiation Laboratory, University of Science and Technology of China, Hefei 230029, China

4 College of Engineering, University of Guelph, Guelph, Ontario N1G 2W1, Canada

*Corresponding author, liumz@cdut.edu.cn (Mingzhe Liu)

## ABSTRACT

*Abstract— Dehazing networks are commonly trained on synthetic hazy–clear pairs, but their performance often drops on real photographs. Synthetic haze generated using the atmospheric scattering model does not fully capture the variability of real haze, and paired real hazy–clear images are scarce. In this work, we propose HazeSpikeMamba, a compact dehazing framework that combines a spiking-inspired local path and an attentive state-space global path in a multi-scale U-Net. The local path uses TPCNNSpike, a new spike-emission scheme inspired by the neighborhood coupling of Pulse-Coupled Neural Network (PCNN). Unlike grouped directional scanning, TPCNNSpike updates all neurons in parallel using the previous firing states of their Gaussian-weighted neighborhoods. The global path adapts the Attentive State-Space Module of MambaIRv2, retaining semantic prompting and sequence reordering while removing the window self-attention branch. Its state-space processing models long-range dependencies with complexity linear in sequence length. For target-domain adaptation, a frozen degradation network, pretrained on paired NH-HAZE data, re-synthesizes haze from the dehazed prediction. The reconstruction error updates only the final restoration layers of HazeSpikeMamba without haze-free labels during adaptation. A shared checkpoint is adapted once on each complete unlabeled target set, making the evaluation dataset-level and transductive rather than zero-shot or per-image optimization. The forward network contains 2.02M active parameters and requires 13.27G nominal MACs (measured with thop at 256×256 input). This adaptation consistently improves BRISQUE and NIMA on RTTS, URHI, and HSTS. On RTTS, BRISQUE decreases from 30.13 to 27.72 and NIMA increases from 4.13 to 4.87. Under this transductive protocol, the adapted model also achieves the best BRISQUE and NIMA on URHI and HSTS among the compared methods.*

Keywords: Image dehazing, spiking neural networks, state-space models, self-supervised fine-tuning, real-world generalization.

## INTRODUCTION

Single-image dehazing aims to recover scene radiance from an observation degraded by atmospheric scattering [1-3]. Haze reduces contrast, obscures distant structures, and changes image color, thereby affecting both visual quality and downstream vision tasks [4-6]. In recent years, supervised deep networks have become the dominant approach to this problem. However, paired hazy–clear images of real scenes are difficult to capture, and most networks are therefore trained with synthetic haze generated by the atmospheric scattering model. Real haze is also affected by spatially varying depth, particle density, illumination, and camera response. These factors are difficult to reproduce during synthesis, causing models trained on synthetic pairs to generalize poorly to real scenes [7].

A practical dehazing model must address three problems. It should be compact enough for deployment, capture both local image details and scene-wide haze distributions, and adapt to real degradation without clear references. Most existing methods address these requirements separately. To address these requirements jointly, we propose a compact multi-scale U-Net that combines a TPCNNSpike-based local path with an attentive state-space global path through stage-wise feature fusion. We further equip the network with a degradation-based self-supervised adaptation scheme to bridge the synthetic-to-real gap.

Existing architectures expose a local–global–efficiency trade-off. Convolutional Neural Network (CNN) dehazing models recover edges and textures effectively but require multi-scale processing, greater depth, or enlarged kernels to cover broad nonuniform haze [8-10]. Transformers model non-local dependencies more directly [11, 12], but full self-attention is expensive at image resolution. Spiking Neural Networks (SNNs) and state-space models offer compact alternatives [13-15], although spatial scan design determines which regions can interact. These limitations motivate the coupled local and global paths developed in this work and reviewed in Section II.

Architecture alone does not remove the synthetic-to-real gap. Unsupervised domain adaptation and unpaired methods incorporate real hazy images through translation, physical priors, or haze re-synthesis [16-18], while self-supervised methods derive targets or consistency signals from the observations themselves [19, 20]. Their effectiveness can depend on hand-crafted priors, estimated depth, adversarial translation, or access to an unpaired clear-image collection. Section II reviews these training strategies and distinguishes them from the setting used here.

HazeSpikeMamba addresses these problems jointly. Its local path replaces the grouped directional accumulation of Orthogonal Leaky-Integrate-and-Fire (OLIF) in DehazeSNN [21] with synchronous Gaussian-neighborhood coupling inspired by Pulse-Coupled Neural Network (PCNN) [22]. Its global path adapts the Attentive State-Space Equation (ASE) and Semantic Guided Neighboring (SGN) mechanisms of MambaIRv2 [23] while removing window Multi-Head Self-Attention (MHSA), because local interaction is already assigned to the spiking-inspired path. For target-domain adaptation, we transfer the reconstruction principle of Low-Res Leads the Way [24] from super-resolution to haze formation: a frozen degradation model maps the dehazed prediction back to the observed hazy domain, so an unlabeled real hazy image supplies its own degradation-consistency signal.

The main contributions of this work are summarized as follows:

1. We propose TPCNNSpike, a lightweight spike-emission mechanism that removes OLIF's directional information-flow constraints and inter-group isolation. Through synchronous Gaussian-neighborhood coupling, it propagates local information in all spatial directions over successive iterations while introducing only two trainable parameters per instance in one channel.

2. We introduce an MHSA-free MambaBlock that captures scene-wide haze distributions through linear-complexity state-space modeling with ASE and SGN. Its global features are fused with the dedicated local path through a stage-level scalar gate after each dual-branch stack, avoiding redundant local modeling through window attention.

3. We develop a degradation-consistency fine-tuning strategy that adapts the model to real haze without paired or haze-free targets during adaptation. A frozen DegModel maps the dehazed prediction back to the observed hazy domain, providing a self-supervised reconstruction signal for updating the final restoration layers.

4. The resulting HazeSpikeMamba model provides a compact yet competitive solution for real-image dehazing, as shown in Fig. 1, with 2.02M parameters and 13.27G MACs. Fine-tuning consistently improves BRISQUE and NIMA on RTTS, URHI, and HSTS.

The remainder of this paper is organized as follows. Section II reviews dehazing architectures and learning from unpaired data. Section III describes the backbone architecture, the two feature paths, and the adaptation scheme. Section IV presents the experimental protocol, benchmark results, and ablation studies. Section V discusses the findings and their limitations, and Section VI concludes the paper.

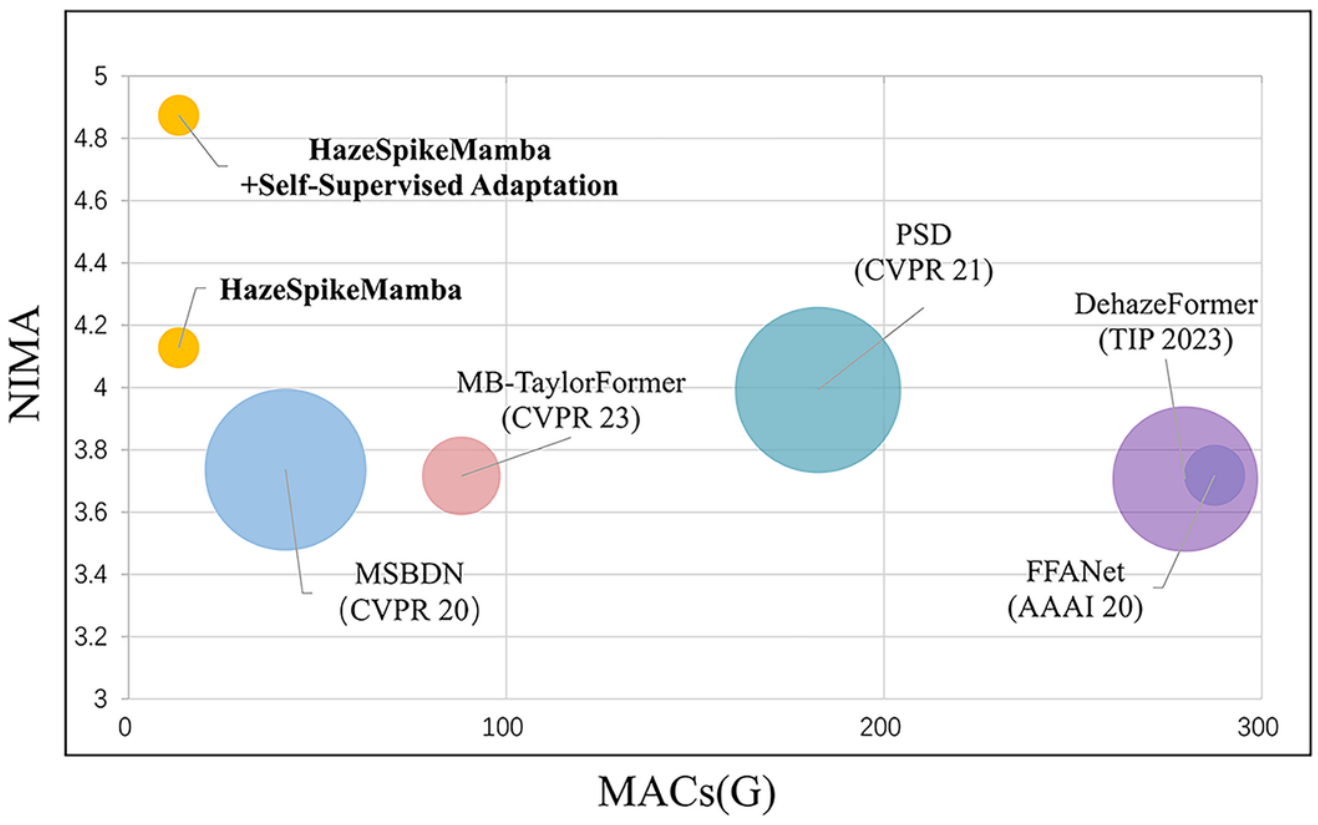

Figure 1. Comparison of computational complexity (MACs), perceptual quality (NIMA), and model size among dehazing methods. Bubble area is proportional to the number of model parameters. HazeSpikeMamba achieves high NIMA with comparatively low reported complexity.

## II. RELATED WORK

### *A. Image dehazing architectures*

Early deep dehazing systems were dominated by CNNs. DehazeNet learns haze-relevant features and estimates transmission locally [10], while MSCNN predicts a coarse transmission map and refines it at a finer scale [8]. Later end-to-end models strengthened feature reuse and scale interaction: FFA-Net combines channel and pixel attention with multi-level feature fusion [25], and MSBDN uses a boosted decoder and dense cross-scale fusion [26]. C2PNet introduces a physics-aware dual-branch unit and curricular contrastive regularization [27], whereas DEA-Net emphasizes local detail through re-parameterized difference convolutions and content-guided attention [28]. These designs improve local restoration and partially address scale variation, but convolution remains spatially local. Covering distant haze regions consequently relies on repeated downsampling, deep stacks, large kernels, or elaborate fusion, which can increase cost and still provide only indirect scene-wide interaction. HazeSpikeMamba retains efficient convolutional detail extraction while injecting multiscale versions of the input and delegates long-range modeling to a separate state-space path.

Transformers address the limited receptive field by using content-adaptive attention [29]. Dehamer combines convolution with a transmission-aware Transformer [30], DehazeFormer develops dehazing-specific normalization and attention within an encoder–decoder [31], and MB-TaylorFormer uses multi-branch, multi-scale embeddings with a Taylor approximation to attention [32]. More recent work continues this hybrid trend: the physics-infused dual-path Transformer couples a physics-aware parameter path with data-driven Transformer restoration [33]. Global attention is well suited to spatially extended haze, but standard Softmax attention scales quadratically with the number of pixels. Windowing limits this cost by restricting interaction, while linear approximations trade exact attention for efficiency. Physics-informed designs can still transfer errors when the assumed scattering model or its estimated parameters do not match real haze. The proposed global path in HazeSpikeMamba instead uses a linear-complexity state-space model and therefore does not require quadratic image-level attention.

SNNs provide another route to compact restoration. DehazeSNN is an early U-Net-like SNN for single-image dehazing and combines multi-scale stages, SKfusion, and OLIF blocks [21]. OLIF performs group-wise horizontal and vertical Leaky Integrate-and-Fire (LIF)-style spatial accumulation, supporting competitive restoration with a small model. However, each update remains directional and confined to a channel group, so positions in different groups do not exchange firing information during that update. Iterative spiking computation may also introduce latency that parameter and MAC counts do not capture. In HazeSpikeMamba, TPCNNSpike directly addresses this spatial-interaction limitation: all units update in parallel from the previous firing states of Gaussian-weighted neighbors, permitting information to propagate in every spatial direction over successive iterations. Because it passes full-precision responses, our method does not claim neuromorphic execution or energy savings.

Recent Mamba-based models seek a better balance between context and complexity. UVM-Net combines convolution with bidirectional state-space blocks in a U-shaped dehazing network [34], and DehazeMamba augments Mamba with quality priors derived from a large multimodal model and channel attention [35]. More recently, CoFiWaveMamba uses a 2D selective-scan Mamba for low-frequency restoration and wavelet/Fourier refinement for directional details [36]. More generally, MambaIRv2 introduces ASE and SGN for image restoration [23]. State-space processing is linear in sequence length, but flattening a two-dimensional image introduces scan-order bias, causal access restrictions, and a need for explicit local-detail modeling. Frequency-decoupled pipelines can recover global structure and texture separately, but errors in coarse low-frequency restoration may propagate to later refinement and the added transforms increase design complexity. HazeSpikeMamba instead uses ASE prompts to supplement causal context and SGN to place semantically related haze regions next to one another in the scan, while TPCNNSpike supplies local detail in parallel at every stage. This pairing also removes MambaIRv2's window-MHSA branch and avoids duplicated local modeling.

### *B. Dehazing without Paired Real-World Supervision*

Because aligned real hazy–clear pairs are scarce, many studies learn from real hazy images without scene-matched clear references. Depending on their data and training protocols, these methods are described as unpaired, unsupervised, self-supervised, semi-supervised, or domain-adaptive. These terms are not interchangeable, but they can overlap. Unpaired methods may use separate collections of hazy and clear images; semi-supervised methods combine paired synthetic data with unlabeled real images; and self-supervised adaptation derives training signals from unlabeled target observations. A domain-adaptation method, for example, may also employ self-supervised losses. We therefore organize the following discussion by the source of supervision: physical priors, cross-domain translation or decomposition, and target-derived consistency or semantic guidance.

Prior-driven methods remove the need for paired real targets by constraining the output directly. Deep DCP trains on real hazy images by minimizing a dark-channel-prior energy [37]. Semi-Supervised Image Dehazing combines supervised losses on synthetic pairs with dark-channel and gradient priors on unlabeled real images [38]. PSD first pretrains a backbone on synthetic pairs and then fine-tunes it on real hazy images using a committee of physical-prior losses [17]. These methods are simple and do not require a learned inverse degradation model, but their supervision inherits the failure modes of the selected priors, particularly in bright, sky, or nonstandard illumination regions.

Translation and decomposition methods instead learn relations between hazy and clear domains. DAD uses bidirectional image translation, two domain-specific dehazing networks, and consistency constraints to bridge synthetic and real data [16]. D4 uses unpaired hazy and clear images, decomposes transmission into density and depth, and re-renders multiple haze levels for self-augmentation [18]. ODCR separates haze-related and haze-unrelated features and applies contrastive constraints in the two spaces [39]. In 2026, OBCOT formulates unpaired dehazing as structure-preserving optimal transport and adapts a one-step Stable Diffusion model with LoRA and domain-specific prompts [40]. These approaches exploit richer distributional information than a single hand-crafted prior, but adversarial translation can be unstable, DAD duplicates domain-specific components, and decomposition or contrastive methods rely on assumptions about depth, haze factorization, and the representativeness of the unpaired clear set. Generative optimal-transport methods further depend on large pretrained priors and optimize distribution-level photo-realism, which does not by itself guarantee scene-specific reconstruction fidelity.

Self-supervised and unsupervised adaptation construct supervision from physical consistency, generated degradations, or pretrained semantic priors. SLAdehazing uses estimated depth and the atmospheric scattering model for self-supervised pretraining, followed by contrastive target adaptation [19]. SSDN jointly predicts the clear image, transmission, and atmospheric light and enforces atmospheric-model and image-prior objectives [20]. BiLaLoRA uses a CLIP-based haze-to-clear text loss and bilevel optimization to select the layers receiving LoRA adapters, thereby reducing the cost of adapting a synthetic-trained model to unlabeled real haze [41]. Such methods can adapt without paired target images, but errors in monocular depth, atmospheric-light estimation, or the assumed scattering model can become errors in the training signal. Semantic text guidance avoids explicit physical estimation but may not sufficiently constrain pixel-level scene fidelity.

Our adaptation stage is closest to synthetic-pretraining followed by self-supervised target fine-tuning, as in PSD and SLAdehazing, but the supervision source is different. Instead of evaluating the dehazed output with fixed priors or an estimated depth map, a frozen learned DegModel re-applies the input's degradation representation and compares the reconstructed haze with the observed haze. No clear image—paired or unpaired—is used during target adaptation, and only the final restoration layers of HazeSpikeMamba are updated. This design avoids direct dependence on dark-channel assumptions and depth estimation, while retaining a compact ordinary forward pass after dataset-level adaptation.

## III. METHODOLOGY

### A. *Overview*

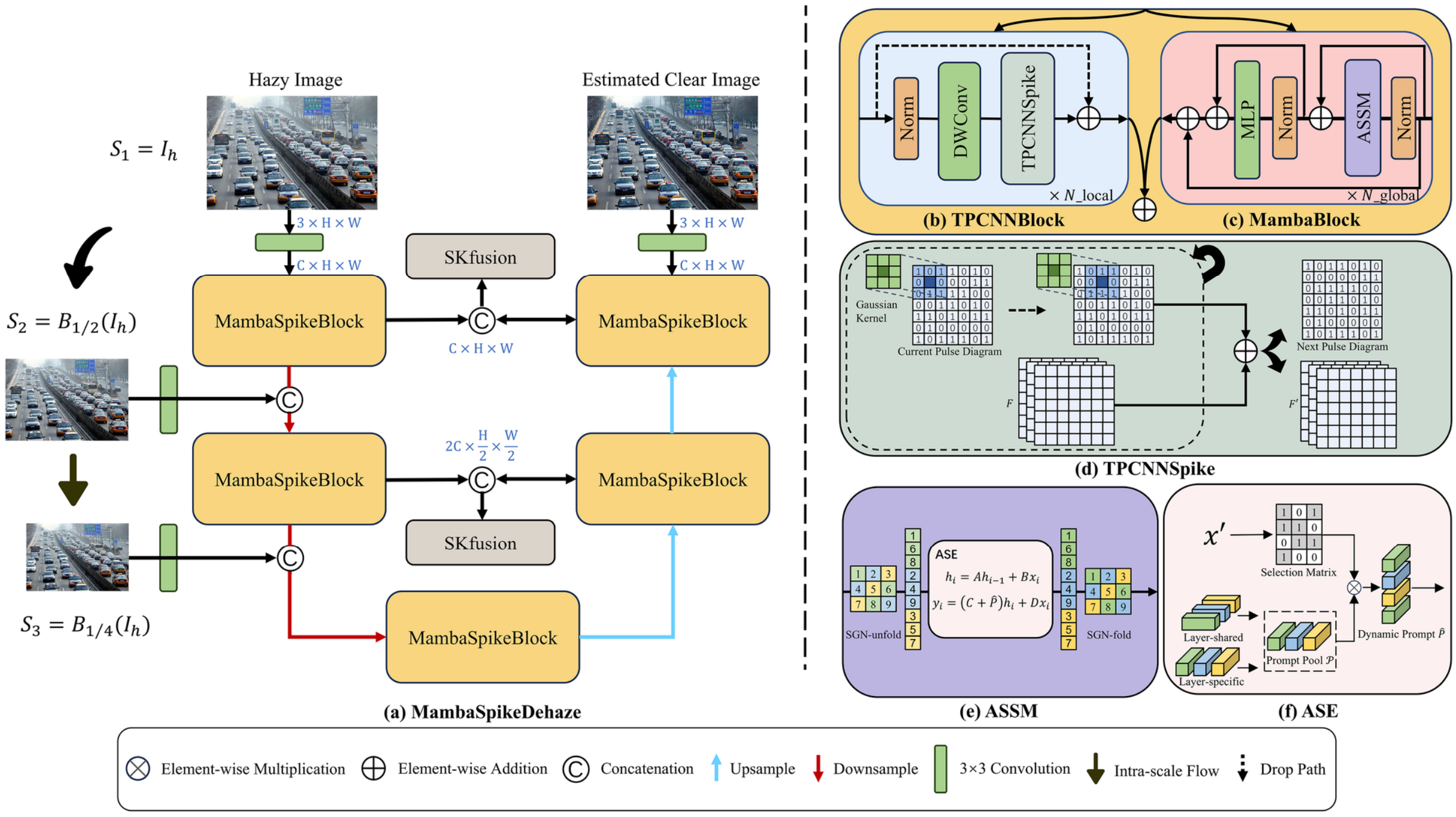

Figure 2. Overall architecture of HazeSpikeMamba. (a) Multi-scale U-shaped HazeSpikeMamba backbone. (b) Local TPCNNBlock. (c) Global MambaBlock. (d) Synchronous Gaussian-neighborhood coupling in TPCNNSpike. (e) Attentive State-Space Module (ASSM) with semantic-guided token reordering. (f) Attentive State-Space Equation (ASE) with dynamic prompt selection. Fig. 2b and 2c show the internal structures of the local and global blocks. In each stage, stacks of these blocks process the same feature map in parallel and are merged once by a stage-level scalar gate.

Let $I_h \in \mathbb{R}^{3\times H\times W}$ denote a hazy image and $I_c$ its latent clear counterpart. The dehazing network $D_\theta$ estimates

$$\hat{I}_c = D_\theta(I_h). \quad (1)$$

HazeSpikeMamba implements $D_\theta$ as a five-stage U-Net [42] with multi-scale image inputs, as shown in Fig. 2a. Each encoder/decoder stage is implemented as a dual-branch MambaSpikeBlock. Given the stage input $F$, the local branch applies a stack of TPCNNBlocks, and the global branch applies a stack of MambaBlocks to the same-resolution feature map in parallel. After both stacks complete, a stage-level scalar gate merges the two branch outputs and adds a residual connection to $F$. The network is first trained on paired synthetic data and then adapted to unlabeled real hazy images using a frozen degradation reconstruction network.

### *B. Multi-scale encoder–decoder*

The apparent scale of haze changes with scene depth. To preserve haze information at different scales, HazeSpikeMamba receives the input image at three resolutions instead of relying only on downsampled backbone features. With $\mathcal{B}$ denoting bilinear resizing,

$$S_1 = I_h, \qquad S_2 = \mathcal{B}_{1/2}(I_h), \qquad S_3 = \mathcal{B}_{1/4}(I_h). \tag{2}$$

A separate $3 \times 3$ convolution projects each scale into feature space. $S_1$ enters the encoder directly; when the encoder reaches the $1/2$ and $1/4$ resolutions, the corresponding scale features are concatenated with the backbone features and fused by a convolution.

The backbone contains three encoder resolutions and two decoder stages. The decoder progressively upsamples the features and merges them with same-scale encoder features through SKfusion [21]. This module adaptively weights the skip and main branches rather than treating both branches equally. The five stages use local and global stacks of depths 8, 8, 16, 8, and 8, with channel dimensions of 12, 24, 48, 24, and 12, respectively. The MLP hidden expansion ratio is 4. Finally, a convolution layer reconstructs the estimated clear image $\hat{I}_c \in \mathbb{R}^{3\times H\times W}$.

### *C. TPCNNBlock and the TPCNNSpike activation*

As shown in Fig. 2b, Trainable PCNN Block (TPCNNBlock) serves as the local feature-extraction path. TPCNNBlock applies residual-pre-normalization, a $1 \times 1$ and depth-wise $3 \times 3$ front-end with normalization and LeakyReLU, TPCNNSpike, and a DropPath residual. The figure shows only the main path for clarity.

The design is motivated by the scanning restriction of OLIF-style grouped activation [21], in which a unit can only use information from earlier units of its own scan group. Following the neighborhood-coupling idea of PCNNs [22], TPCNNSpike updates all units synchronously: at every iteration, each unit reads the previous-iteration firing states of its spatial neighborhood, so that information propagates outward in all directions as iterations accumulate. In this way, the temporal membrane accumulation of a conventional SNN is converted into accumulation over iteration rounds.

For location $(i,j)$ at iteration $t$, let $x_{ij}$ be the input stimulus, $u_{ij}^t$ the membrane potential, and $o_{ij}^t \in \{0,1\}$ the firing indicator. The update is defined as

$$y_{ij}^{t+1} = x_{ij}, \qquad r_{ij}^0 = x_{ij}, \tag{3}$$

$$L_{ij}^{t+1} = \sum\nolimits_{k,l} M_{ijkl} \cdot o_{kl}^t, \tag{4}$$

$$u_{ij}^{t+1} = \tau u_{ij}^t\left(1-o_{ij}^t\right) + y_{ij}^{t+1}\left(1+L_{ij}^{t+1}\right), \tag{5}$$

$$o_{ij}^{t+1} = \begin{cases} 1 & u_{ij}^{t+1} > V_{\text{th}} \\ 0 & u_{ij}^{t+1} \le V_{\text{th}} \end{cases}, \tag{6}$$

$$r_{ij}^{t+1} = o_{ij}^{t+1} ReLU\left(u_{ij}^{t+1} - V_{\text{th}}\right) + r_{ij}^t, \tag{7}$$

where $M_{ijkl}$ is a Gaussian coupling kernel that weights the influence of neuron $(k,l)$ on neuron $(i,j)$ according to their spatial distance. The decay parameter $\tau$ controls the membrane potential carried between iterations, and $V_{\text{th}}$ is the firing threshold. Below this threshold, the membrane potential accumulates through decay and input stimulation. Once the threshold is exceeded, the neuron fires. The binary firing state $o_{ij}^{t+1}$ is used only within the iterative coupling: it enters the reset term in Eq. (5) and the neighborhood interaction in Eq. (4). Separately, a full-precision response $r_{ij}^{t+1}$ accumulates the above-threshold surplus over successive iterations. After the final iteration, $r$ is forwarded as the TPCNNSpike output to the next network layer. Only $\tau$ and $V_{\text{th}}$ are trainable within the activation: each TPCNNSpike instance learns one decay coefficient and one threshold per channel. Because the forwarded feature remains full precision rather than a binary spike train, the module can be trained by ordinary gradient descent.

### *D. MambaBlock: an MHSA-free attentive state-space path*

TPCNNBlock models local interactions, but it does not provide scene-wide context when haze is dense or spatially extended. We therefore use MambaBlock as the global path. This block is based on the Attentive State-Space Module (ASSM) of MambaIRv2 [23]. As shown in Fig. 2c, MambaBlock contains normalization, ASSM, a residual connection, an MLP, and a second residual connection. The window-MHSA branch of the original MambaIRv2 block is removed because local interaction is already handled by TPCNNBlock.

Given a feature map $X \in \mathbb{R}^{C\times H\times W}$, ASSM first applies positional encoding and flattens the map into a sequence $X' \in \mathbb{R}^{L\times C}$ with $L = H \times W$. A standard state-space model processes this sequence causally as

$$h_i = Ah_{i-1} + Bx_i, \qquad (8)$$

$$y_i = Ch_i + Dx_i. \qquad (9)$$

In an image, all pixels are available simultaneously, whereas the standard state-space equation only accesses preceding tokens. ASE reduces this restriction by adding a position-dependent semantic prompt to the output projection:

$$y_i = \left(C + \hat{P}_i\right)h_i + Dx_i. \qquad (10)$$

The prompts come from a pool $P \in \mathbb{R}^{T\times D}$, where $T$ is the number of prompts and $D$ the hidden state dimension. The pool is factorized as

$$P = M \times N, \quad M \in \mathbb{R}^{T\times r}, \quad N \in \mathbb{R}^{r\times D}, \quad r \ll \min(T, D), \qquad (11)$$

where $N$ is shared across blocks and $M$ is block specific. This factorization provides a shared semantic basis while allowing each block to learn its own combinations of that basis. Given the flattened token features, the semantic router projects each token from $C$ channels to $T$ category scores and applies LogSoftmax to obtain log-probabilities over the prompt pool. Hard Gumbel–Softmax then produces a one-hot routing matrix $R \in \mathbb{R}^{L\times T}$, and the token-specific prompts are obtained as $\hat{P} = RP \in \mathbb{R}^{L\times D}$. Adding $\hat{P}_i$ to the output projection $C$ in the ASE equation above supplies learned category-level context intended to alleviate the lack of access to unscanned tokens in a one-directional state-space scan.

The same routing matrix also determines the SGN permutation. Before the state-space scan, SGN groups tokens assigned to the same prompt category and concatenates the resulting groups in category order. Consequently, tokens that are distant in the original image but assigned to the same learned category become closer in the one-dimensional sequence. ASE processes this reordered sequence, after which the inverse permutation restores every token to its original spatial position and a token-wise linear projection produces the block output. SGN therefore shortens the sequence distance between similarly routed tokens and facilitates their interaction during the scan without introducing quadratic self-attention. The routing categories are learned end-to-end and are not explicitly supervised as haze-density or scene-depth labels.

### *E. Gated local–global fusion*

Within each stage, the local and global branches receive the same input feature map $F$. The TPCNN stack output $F_{loc}$ and the Mamba stack output $F_{glb}$ are merged by a learned scalar gate $\lambda$, denoted as

$$F_{\text{out}} = \lambda F_{loc} + (1-\lambda)F_{glb} + F. \qquad (12)$$

Thus, fusion is performed once per stage and is shared across all channels and spatial locations. The five stages therefore contain five scalar gate parameters in total. TPCNNBlock mainly provides local edges and textures, whereas MambaBlock provides scene-level context; their individual and combined effects are evaluated in the component ablation in Section IV.

## *F. Haze degradation reconstruction*

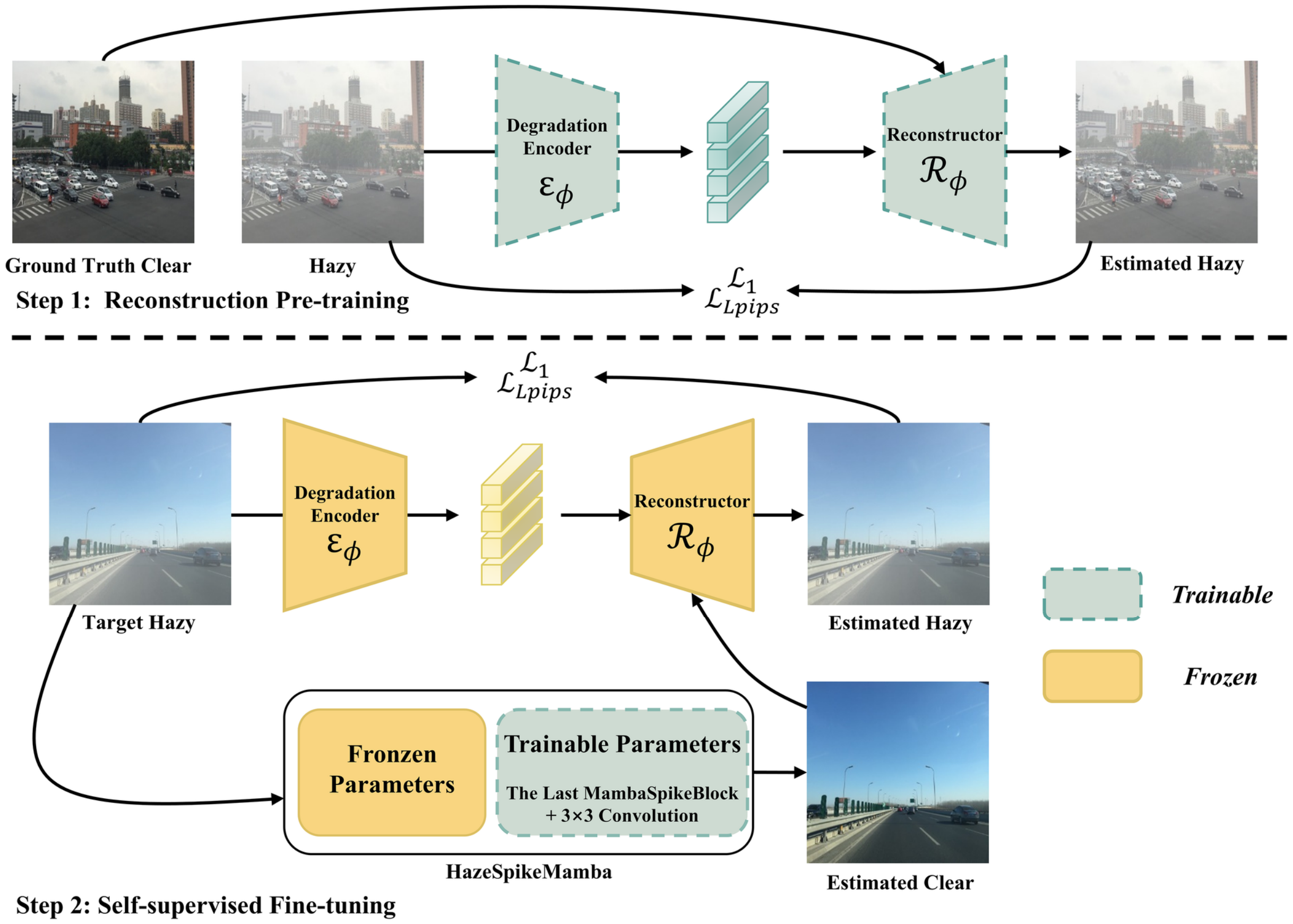

Figure 3. Two-stage degradation-consistency adaptation framework. In Step 1, DegModel learns to reconstruct a hazy image from its clear counterpart and degradation representation using L1 and LPIPS losses. In Step 2, DegModel is frozen and re-applies the degradation extracted from an unlabeled target hazy image to the HazeSpikeMamba output. The reconstruction discrepancy provides self-supervision for fine-tuning selected HazeSpikeMamba parameters without clear target images. Dashed teal and solid yellow modules denote trainable and frozen parameters, respectively.

The adaptation scheme uses the forward haze process to supervise the dehazing network. Compared with recovering an unknown clear image, applying an estimated degradation to an image is a more constrained problem. The atmospheric scattering model describes a hazy observation as

$$I_h(x) = I_c(x)t(x) + A\big(1 - t(x)\big), \tag{13}$$

where $t(x)$ is the transmission and $A$ is the atmospheric light. Haze formation combines attenuation with airlight and can be represented as a forward map conditioned on a clear image and a haze description. We let a degradation model, namely DegModel, learn this map with an encoder–generator architecture. As shown in Fig. 3, its encoder $\varepsilon_\phi$ extracts a degradation representation from a hazy image, and its generator $\mathcal{R}_\phi$ uses this representation to apply the degradation to a clear image:

$$g = \varepsilon_\phi(I_h), \quad \widehat{I_h} = \mathcal{R}_\phi(I_c, g). \tag{14}$$

Once trained, DegModel is frozen throughout self-supervised adaptation.

## *G. Self-supervised fine-tuning*

At adaptation time, as shown in Fig. 3, an unlabeled real hazy image is first dehazed by the pretrained HazeSpikeMamba:

$$\hat{I}_c = D_\theta(I_h). \quad (15)$$

The frozen degradation model then re-synthesizes the haze from the prediction and the input's own degradation representation:

$$\widehat{I_h} = \mathcal{R}_\phi(\hat{I}_c, \varepsilon_\phi(I_h)). \quad (16)$$

If $\hat{I}_c$ is a faithful clear version of the scene, re-applying the observed degradation should reproduce the observed haze. Consequently, the discrepancy between $\tilde{I}_h$ and $I_h$ serves as a supervision signal that never references a haze-free target. The fine-tuning objective is the unweighted sum of a reconstruction term and a Learned Perceptual Image Patch Similarity (LPIPS) term [43]:

$$\mathcal{L}_{\text{ssl}} = \left\|\widehat{I_h} - I_h\right\|_1 + \mathcal{L}_{\text{LPIPS}}(\tilde{I}_h, I_h). \quad (17)$$

Adaptation is performed once at the dataset level rather than independently for each image. During this process, DegModel remains frozen and only a restricted subset of the restoration network is updated. The resulting shared model subsequently processes every image by ordinary forward inference without per-image optimization. The complete optimization and checkpoint-selection protocol is provided in Section IV.

This objective imposes degradation consistency, but it does not form a complete reconstruction cycle. A low loss indicates that the predicted clear image can reproduce the observation under the learned degradation model. It does not show that the clear image is uniquely determined. Freezing DegModel and updating only the final restoration layers of HazeSpikeMamba reduce the possibility of a physically implausible solution, but they cannot eliminate it. This limitation is discussed further in Section V.

## IV. EXPERIMENT

### *A. Experimental setup*

All models were implemented in PyTorch and trained on a single NVIDIA GeForce RTX 4090 using AdamW with $\beta_1 = 0.9$ and $\beta_2 = 0.999$. Training used $256 \times 256$ image patches. The baseline HazeSpikeMamba was trained with a batch size of 2 on O_RESIDE_6K, a subset of RESIDE-6K [44] containing 3,000 synthetic outdoor images. Outdoor scenes were selected because real haze occurs predominantly in outdoor environments; in preliminary experiments, this subset also yielded better generalization to real scenes than training on NH-HAZE. The training objective was the unweighted sum of L1 and LPIPS losses. The learning rates were initialized to $5 \times 10^{-5}$ for $\tau$ and $V_{\text{th}}$ and to $10^{-4}$ for all remaining parameters, and both were decayed to $10^{-6}$. The two TPCNNSpike parameters were initialized to 0.7, and the network used ten iterations.

DegModel was trained with a batch size of 16 on NH-HAZE [45], which contains nonhomogeneous haze generated physically with haze machines. Its optimization used the unweighted sum of L1 and LPIPS losses and a learning-rate schedule from $10^{-4}$ to $10^{-6}$. Peak signal-to-noise ratio (PSNR) was used for checkpoint selection in both training stages. Dehazing checkpoints were selected on paired RESIDE validation images, whereas DegModel checkpoints were selected by comparing reconstructed hazy images with their original hazy inputs. No clear image from a real target domain was used for checkpoint selection. The influence of the DegModel training source is examined in the ablation study.

For each target benchmark, self-supervised adaptation started from the synthetic-trained dehazing weights and was performed once on the complete test set for 10 epochs. A batch size of 1 was used, with one random $256 \times 256$ patch sampled at each step; a dataset of $N$ images therefore produced approximately $10N$ gradient updates. Preliminary comparisons of different trainable-layer configurations showed that updating only the final MambaSpikeBlock and the unembedding layer produced the best result, and all remaining restoration layers and DegModel were consequently frozen. This configuration preserves the synthetic-domain features learned by the early layers while adapting the final restoration stages to the real domain. Because larger learning rates caused unstable optimization and overfitting to the reconstruction objective, the adaptation learning rate was set to $10^{-9}$. Only hazy target images contributed to the adaptation loss, and no target-domain clear images were used for either optimization or model selection. The "+ self-supervised adaptation" results report the shared checkpoint obtained after this dataset-level adaptation, and inference required no further optimization.

Evaluation was conducted on three real-image datasets from the RESIDE benchmark, none of which provides paired clear references. RTTS contains 4,322 real hazy images from the Real-world Task-driven Testing Set, URHI contains 4,810 real hazy images, and HSTS contains 10 high-resolution real hazy images. In the absence of reference images, performance was assessed using BRISQUE [46] and SSEQ [47], for which lower values indicate better quality, and NIMA [48], for which higher values are preferred. The comparison includes classical prior-based approaches, CNNs, Transformers, and recent deep models: DCP, MSCNN, DehazeNet, AODNet, MSBDN, DAD, FFANet, PSD, D4, Dehamer, MB-TaylorFormer, C2PNet, DehazeFormer, and DEANet. The deployed HazeSpikeMamba network contains 2.02M active forward-path parameters and requires 13.27G MACs, as measured with thop at an input resolution of $256 \times 256$. Parameter counts and MACs reported for competing methods were taken from their respective publications and may have been measured at different resolutions or with different tools; they therefore provide approximate model-scale comparisons rather than hardware-matched runtime measurements.

### *B. Quantitative analysis*

The three benchmarks test the same synthetic-trained model under different real-image distributions and resolutions. We first examine each dataset separately and then analyze the cross-dataset pattern. Importantly, the un-adapted and adapted results answer different questions: the former measures direct transfer from synthetic training, whereas the latter measures dataset-level transductive adaptation using the unlabeled hazy target set. Fine-tuning changes the weights but not the deployed architecture, so both variants retain the same 2.02M active parameters and 13.27G forward MACs.

The RTTS evaluation uses BRISQUE, NIMA, and SSEQ to assess complementary aspects of perceptual quality and determine whether adaptation produces consistent improvements across the three measures.

TABLE I QUANTITATIVE COMPARISON ON RTTS

| Methods | | BRISQUE↓ | NIMA↑ | SSEQ↓ | #Param | MACs |
|---|---|---|---|---|---|---|
| DCP [49] | TPAMI 11 | 38.7186 | 3.1589 | 40.3328 | - | - |
| MSCNN [8] | ECCV 16 | 33.0982 | 3.7045 | 39.0049 | - | - |
| DehazeNet [10] | TIP 16 | 35.0086 | 3.7163 | 41.0284 | 0.009 M | 0.581 G |
| AODNet [50] | ICCV 17 | 32.0727 | 3.8782 | 35.8573 | - | - |
| MSBDN [26] | CVPR 20 | 27.9514 | 3.7364 | **35.0615** | 31.35 M | 41.54 G |
| DAD [16] | CVPR 20 | 32.4602 | 3.2524 | 35.9945 | - | - |
| FFANet [25] | AAAI 20 | 33.2453 | 3.7183 | 37.3002 | 4.46 M | 287.5 G |
| PSD [17] | CVPR 21 | <u>27.5818</u> | 3.9922 | 36.0744 | 33.11 M | 182.5 G |
| D4 [18] | CVPR 22 | 33.2184 | 3.7232 | 39.6694 | 10.7 M | 2.246 G |
| Dehamer [30] | CVPR 22 | 33.8739 | 3.733 | 38.1866 | 132.50 M | 48.93 G |
| MB-TaylorFormer [32] | CVPR 23 | 33.1165 | 3.7163 | 37.6693 | 7.43 M | 88.1 G |
| C2PNet [27] | CVPR 23 | 34.2688 | 3.7146 | 38.0314 | 7.17 M | - |
| DehazeFormer [31] | TIP 23 | 32.6476 | 3.7061 | 37.9226 | 25.44 M | 279.7 G |
| DEANet [28] | TIP 24 | 30.4837 | 3.7856 | 37.2266 | 3.65 M | 32.23 G |

| | | | | |
|---|---|---|---|---|
| HazeSpikeMamba | 30.1253 | 4.1280 | 36.2194 | 2.02 M | 13.27 G |
| HazeSpikeMamba + self-supervised adaptation | **27.7184**[a] | **4.8739** | 35.3783[a] | | |

a. The best results are highlighted in bold, while the second-best results are underlined.

As shown in Table I, HazeSpikeMamba achieves the highest NIMA even before adaptation, scoring 4.1280 compared with 3.9922 for PSD, the best-performing competing method. This suggests that the compact local-global backbone generalizes well to RTTS despite not being trained on it. However, its BRISQUE and SSEQ scores are not the best, so this advantage does not extend to all no-reference metrics. After fine-tuning, BRISQUE decreases from 30.1253 to 27.7184, NIMA increases from 4.1280 to 4.8739, and SSEQ decreases from 36.2194 to 35.3783. Because all three metrics move in the favorable direction, the NIMA gain does not come at the expense of perceptual quality measured by BRISQUE and SSEQ.

The adapted model ranks first in NIMA and second in both BRISQUE and SSEQ. It trails PSD by only 0.1366 in BRISQUE and MSBDN by 0.3168 in SSEQ, while exceeding PSD, the strongest non-HazeSpikeMamba competitor in NIMA, by 0.8817. Thus, adaptation improves NIMA most clearly, while keeping BRISQUE and SSEQ close to the best reported results. The model is also considerably smaller: PSD and MSBDN have 16.4 and 15.5 times as many parameters, respectively.

TABLE II QUANTITATIVE COMPARISON ON URHI

| Methods | | BRISQUE↓ | NIMA↑ | #Param | MACs |
|---|---|---|---|---|---|
| DCP [49] | TPAMI 11 | 35.6648 | 3.1950 | - | - |
| MSCNN [8] | ECCV 16 | 33.3838 | 3.6505 | - | - |
| DehazeNet [10] | TIP 16 | 33.2972 | 3.6364 | 0.009 M | 0.581 G |
| AODNet [50] | ICCV 17 | 29.6924 | 3.9057 | - | - |
| MSBDN [26] | CVPR 20 | 23.1545 | 3.7418 | 31.35 M | 41.54 G |
| DAD [16] | CVPR 20 | 29.8898 | 3.077 | - | - |
| FFANet [25] | AAAI 20 | 28.4541 | 3.6938 | 4.46 M | 287.5 G |
| PSD [17] | CVPR 21 | 32.2545 | 3.8416 | 33.11 M | 182.5 G |
| D4 [18] | CVPR 22 | 29.7474 | 3.7474 | 10.7 M | 2.246 G |
| Dehamer [30] | CVPR 22 | 28.9035 | 3.7089 | 132.50 M | 48.93 G |
| MB-TaylorFormer [32] | CVPR 23 | 29.0139 | 3.6828 | 7.43 M | 88.1 G |
| C2PNet [27] | CVPR 23 | 31.59 | 3.6542 | 7.17 M | - |
| DehazeFormer [31] | TIP 23 | 29.5789 | 3.6785 | 25.44 M | 279.7 G |
| DEANet [28] | TIP 24 | 25.7643 | 3.7585 | 3.65 M | 32.23 G |
| HazeSpikeMamba | | 24.2273 | 4.3781[a] | 2.02 M | 13.27 G |
| HazeSpikeMamba + self-supervised adaptation | | **22.4291**[a] | **5.0422** | | |

a. The best results are highlighted in bold, while the second-best results are underlined.

The URHI results test whether the same behavior persists on a second, larger collection of real hazy images. As shown in Table II, the un-adapted model is already competitive on URHI. Its BRISQUE score of 24.2273 ranks second among the pretrained methods, behind only MSBDN, while its NIMA score of 4.3781 is the highest overall. Direct transfer from synthetic training data is therefore strong on both metrics. After adaptation, BRISQUE decreases by 1.7982 to 22.4291, and NIMA increases by 0.6641 to 5.0422, placing the model first on both metrics. It outperforms MSBDN by 0.7254 in BRISQUE and AODNet, the strongest competing method in NIMA, by 1.1365.

These results show that adaptation builds on an already strong NIMA score rather than compensating for weak direct transfer. It also moves the model from second to first in BRISQUE, suggesting that the degradation-consistency objective helps account for target-domain statistics not captured during synthetic pretraining.

TABLE III QUANTITATIVE COMPARISON ON HSTS

| Methods | | BRISQUE↓ | NIMA↑ | #Param | MACs |
|---|---|---|---|---|---|
| DCP [49] | TPAMI 11 | 36.5323 | 2.6676 | - | - |
| MSCNN [8] | ECCV 16 | 32.9732 | 3.1166 | - | - |
| DehazeNet [10] | TIP 16 | 33.4765 | 3.2102 | 0.009 M | 0.581 G |
| AODNet [50] | ICCV 17 | 31.1289 | 3.1952 | - | - |
| MSBDN [26] | CVPR 20 | 40.0361 | 2.5447 | 31.35 M | 41.54 G |
| DAD [16] | CVPR 20 | 33.6936 | 2.7378 | - | - |
| FFANet [25] | AAAI 20 | 28.4745 | 2.9883 | 4.46 M | 287.5 G |
| PSD [17] | CVPR 21 | 33.2621 | 3.3282 | 33.11 M | 182.5 G |
| D4 [18] | CVPR 22 | 30.8643 | 3.1189 | 10.7 M | 2.246 G |
| Dehamer [30] | CVPR 22 | 28.3641 | 2.9923 | 132.50 M | 48.93 G |
| MB-TaylorFormer [32] | CVPR 23 | 40.0805 | 2.5054 | 7.43 M | 88.1 G |
| C2PNet [27] | CVPR 23 | 28.6815 | 3.0059 | 7.17 M | - |
| DehazeFormer [31] | TIP 23 | 29.0388 | 3.0043 | 25.44 M | 279.7 G |
| DEANet [28] | TIP 24 | 28.0282 | 2.9779 | 3.65 M | 32.23 G |
| HazeSpikeMamba | | <u>26.2231</u>[a] | <u>4.4323</u> | 2.02 M | 13.27 G |
| HazeSpikeMamba + self-supervised adaptation | | **23.8019**[a] | **5.4706** | | |

a. The best results are highlighted in bold, while the second-best results are underlined.

On HSTS, HazeSpikeMamba outperforms all competing methods on both metrics even before adaptation, as presented in Table III. Its BRISQUE score is 1.8051 lower than that of DEANet, while its NIMA score is 1.1041 higher than that of PSD. After fine-tuning, BRISQUE decreases from 26.2231 to 23.8019, and NIMA increases from 4.4323 to 5.4706. These are the largest adaptation gains across the three datasets, with improvements of 2.4212 in BRISQUE and 1.0383 in NIMA. The strong HSTS performance suggests that the model handles haze at different spatial extents effectively.

Several trends are consistent across the three datasets. First, the un-adapted model achieves the highest NIMA in every table, showing that the backbone retains strong perceptual performance under direct synthetic-to-real transfer. Second, adaptation improves every reported metric. BRISQUE decreases by 1.7982 to 2.4212, NIMA increases by 0.6641 to 1.0383, and SSEQ decreases by 0.8411 on RTTS. The benefit of adaptation is therefore not limited to a single dataset or metric. Third, the model remains considerably smaller than most recent deep baselines. For example, DehazeFormer uses 12.6 times as many parameters and approximately 21.1 times the reported MACs. These results indicate a favorable balance between perceptual quality and model complexity.

These rankings reflect no-reference perceptual quality rather than fidelity to an unavailable clear image. BRISQUE and NIMA consistently favor adaptation across all three datasets, with SSEQ providing additional support on RTTS. However, these metrics cannot determine the accuracy of scene radiance or color recovery. The next section therefore complements these results with a qualitative evaluation.

### C. *Qualitative analysis*

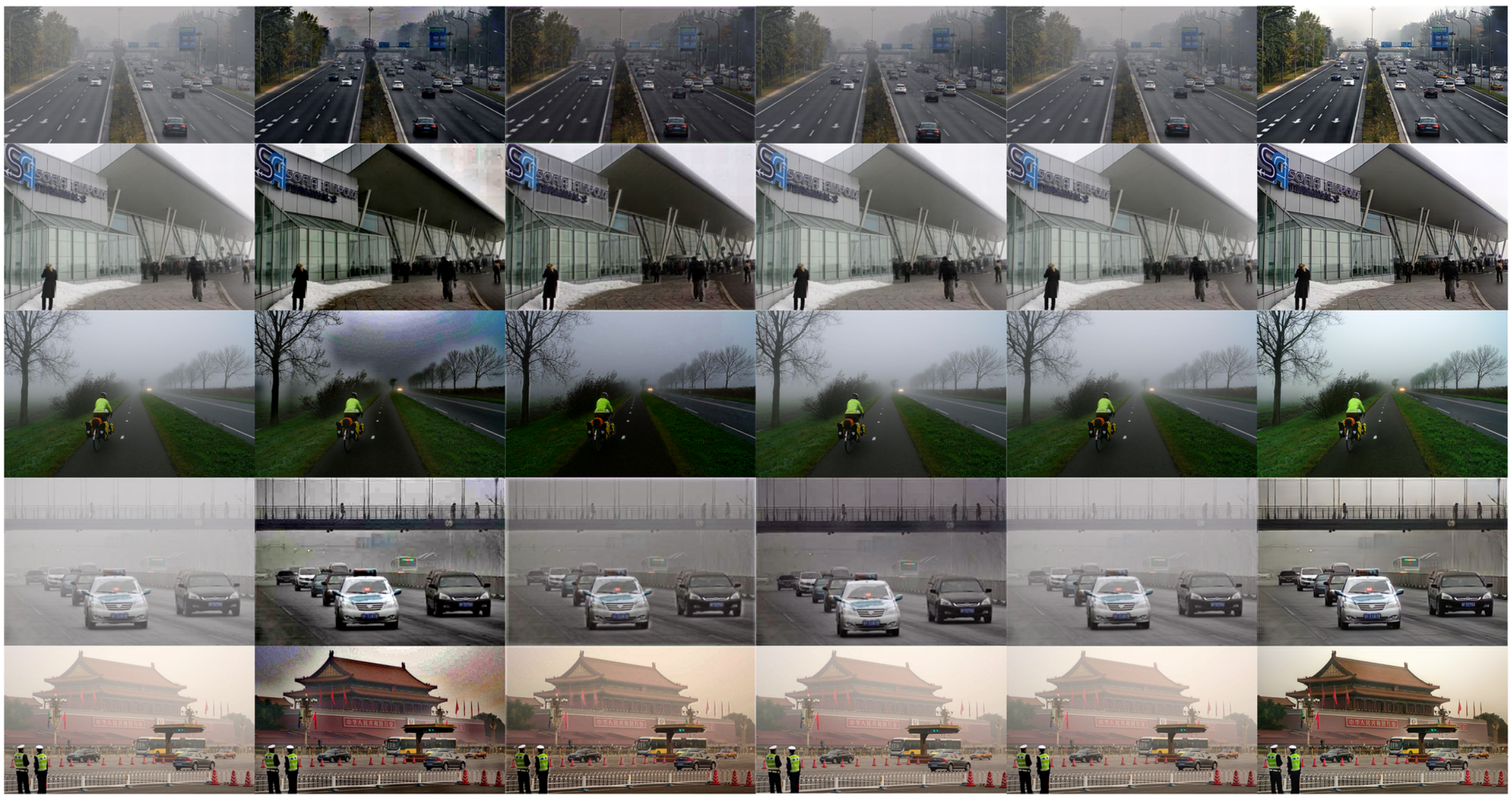

(a) Hazy (b) DCP (c) AODNet (d) C2PNet (e) HazeSpikeMamba (f) HazeSpikeMamba + Self-Supervised Adaptation

Figure 4. Qualitative comparison on five real-world hazy scenes from the RTTS dataset. From left to right: hazy input, DCP, AODNet, C2PNet, HazeSpikeMamba, and HazeSpikeMamba with self-supervised adaptation. The adapted model improves distant visibility and color recovery while preserving natural exposure and clean foreground edges.

Fig. 4 compares the methods on five RTTS scenes. Without adaptation, HazeSpikeMamba removes much of the haze while keeping the overall brightness and color close to the input scene. Unlike DCP, it does not over-darken the road or vegetation, and it introduces no obvious halos along vehicle outlines, building edges, or pedestrians. Some haze nevertheless remains in distant regions, particularly in the sky. Fine-tuning mainly improves these difficult areas: the sky becomes clearer in the first row, distant structures are easier to distinguish, and the lawn in the third row recovers a more natural green. Foreground edges remain clean after adaptation. Thus, the main benefit of fine-tuning is improved background visibility and color correction without sacrificing the stable, artifact-free foreground reconstruction of the original model.

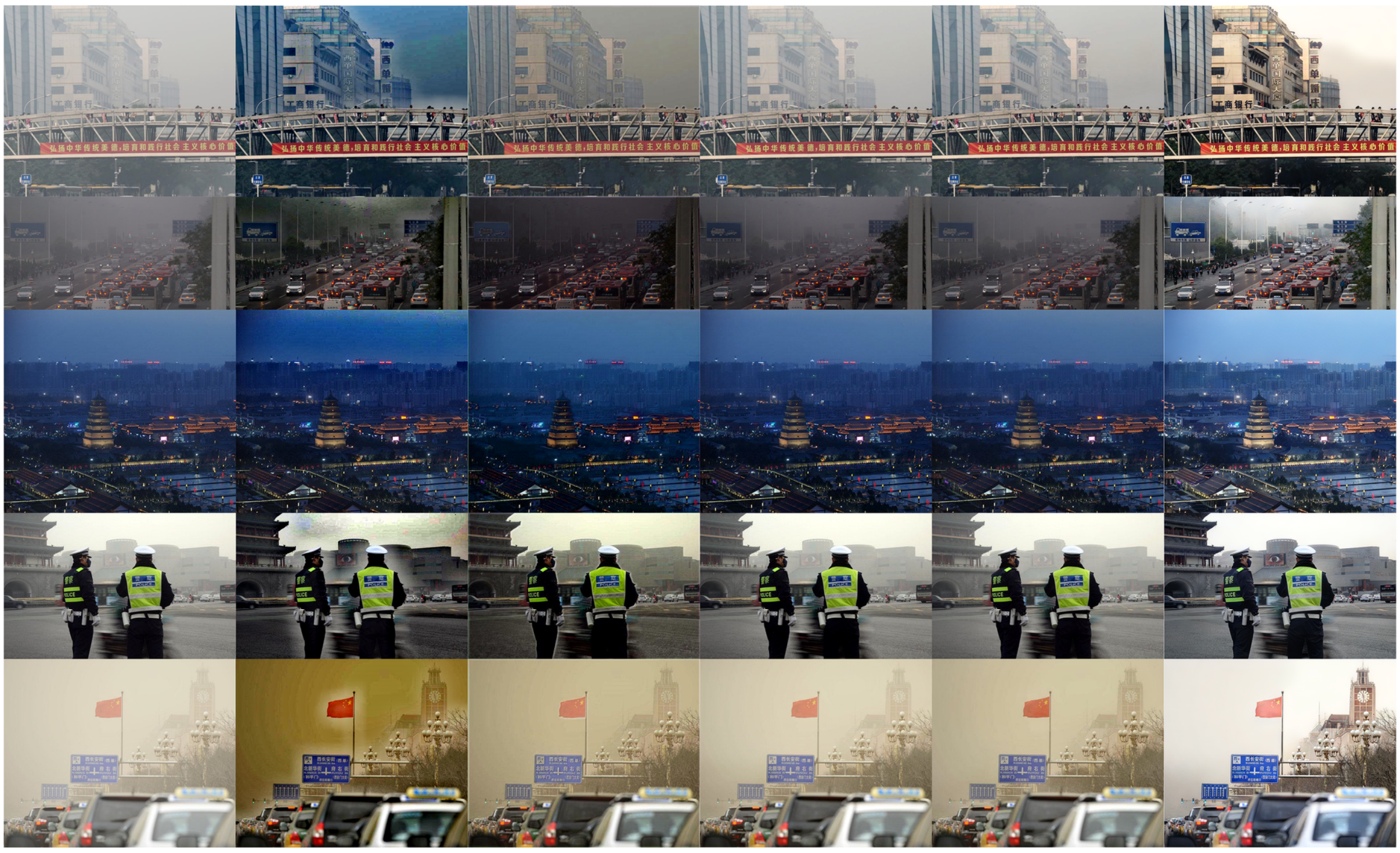

(a) Hazy (b) DCP (c) AODNet (d) C2PNet (e) HazeSpikeMamba (f) HazeSpikeMamba + Self-Supervised Adaptation

Figure 5. Qualitative comparison on five real-world hazy scenes from the URHI dataset. From left to right: hazy input, DCP, AODNet, C2PNet, HazeSpikeMamba, and HazeSpikeMamba with self-supervised adaptation. Adaptation improves distant-scene visibility and reduces color casts while preserving natural exposure and clean object boundaries.

Fig. 5 highlights the effect of adaptation on URHI. Before adaptation, HazeSpikeMamba retains a natural exposure and clean object boundaries, avoiding the strong darkening produced by AODNet, although some background haze remains. Adaptation reveals more of the building façades in the first row and improves the visibility of distant traffic and road signs in the second. The difference is clearest in the last row: the yellow cast is reduced, the clock tower becomes distinct, and the edges of the flag and vehicles remain clean. The night scene also retains its blue tone and point-light sources rather than being uniformly brightened.

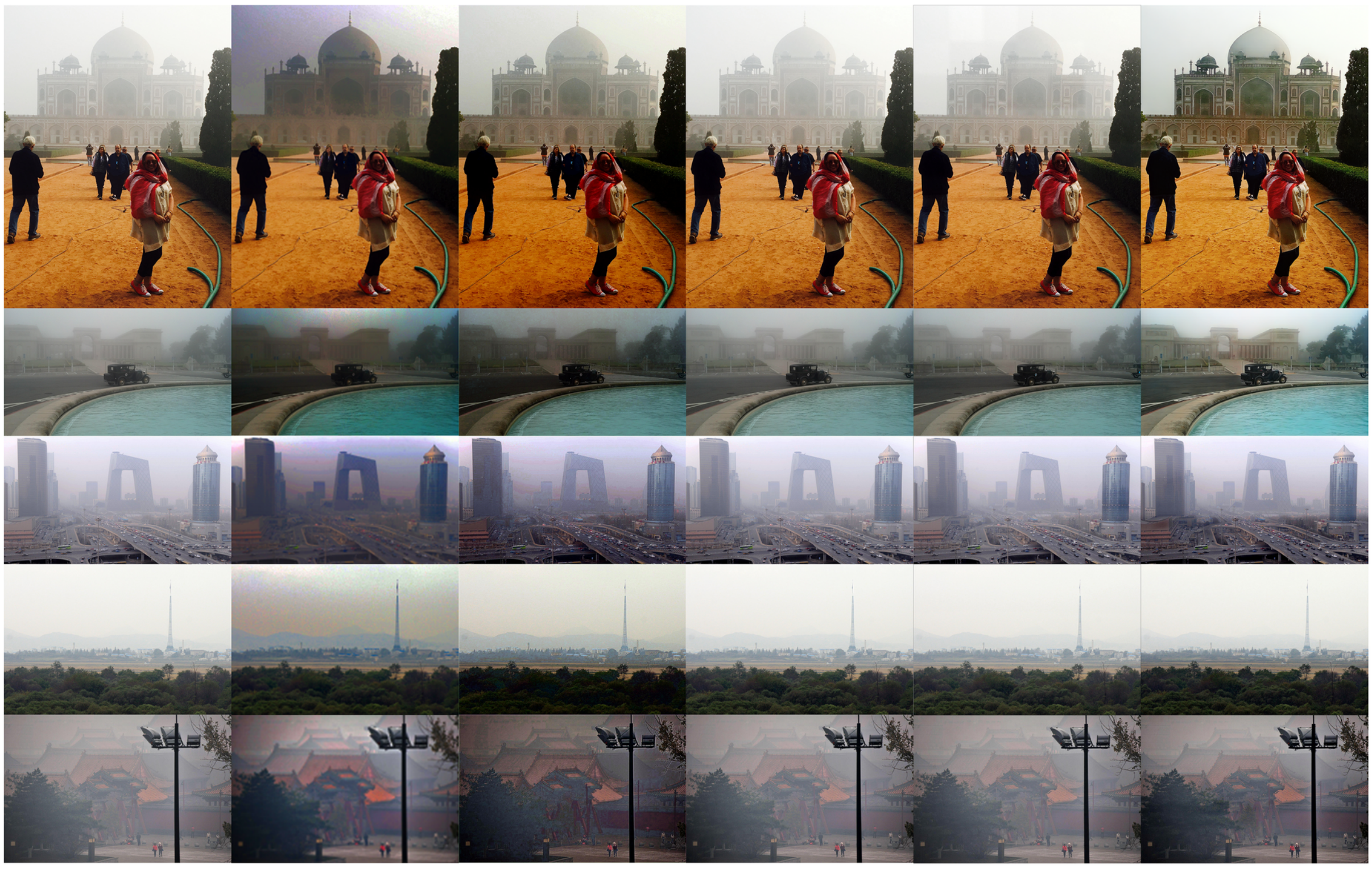

(a) Hazy (b) DCP (c) AODNet (d) C2PNet (e) HazeSpikeMamba (f) HazeSpikeMamba + Self-Supervised Adaptation

Figure 6. Qualitative comparison on five high-resolution hazy scenes from the HSTS dataset. From left to right: hazy input, DCP, AODNet, C2PNet, HazeSpikeMamba, and HazeSpikeMamba with self-supervised adaptation. The adapted model improves long-range visibility and structural detail while avoiding excessive darkening, color shifts, and halo artifacts.

Fig. 6 extends the comparison to the higher-resolution HSTS images. The effect of adaptation is most evident in the first and third rows: the architectural details of the Taj Mahal become visible, and much more of the distant city layout can be seen. The improvement is more restrained in the fourth and fifth rows, but the tree edges, skyline, and overlapping rooflines are still better separated. DCP and AODNet recover contrast by markedly darkening some scenes and shifting their colors, as seen in the orange ground in the first row and the cyan pool in the second. HazeSpikeMamba avoids these large changes in appearance while retaining clear boundaries around people, vegetation, and buildings. Its adapted output therefore provides better long-range visibility without obvious halos or excessive contrast.

### *D. Component ablation*

We assess the contributions of the local TPCNN and global state-space pathways by replacing one or both specialized blocks with MLP modules while retaining the overall U-Net framework. Starting from the MLP baseline, TPCNNBlock and MambaBlock are introduced separately and then jointly in the complete model. All variants are trained on O_RESIDE_6K and evaluated on URHI without test-time adaptation.

TABLE IV ABLATION OF THE LOCAL AND GLOBAL FEATURE PATHWAYS

| Models | BRISQUE↓ | NIMA↑ | #Param | MACs |
|---|---|---|---|---|
| Baseline | 33.227 | 3.45 | 1.37M | 17.11G |
| +TPCNNBlock | 25.906 | 4.016 | 1.14M | 12.4G |
| +MambaBlock | 27.564 | 3.69 | 0.43M | 5.41G |

| Models | BRISQUE↓ | NIMA↑ | #Param | MACs |
|---|---|---|---|---|
| HazeSpikeMamba | **24.2273**[a] | **4.3781** | 2.02M | 13.27G |

a. The best results are highlighted in bold.

Table IV reveals two main trends. First, the TPCNN-only variant performs better than the Mamba-only variant on both perceptual metrics, suggesting that local neighborhood interactions play the more prominent role in this setting. This result is consistent with the spatially varying nature of haze removal, for which recovering edges and fine textures requires effective exchange of information among adjacent features. The global pathway is nevertheless not redundant: the complete model surpasses both single-path variants, indicating that long-range context provides information that cannot be recovered through local interactions alone. The two pathways therefore appear to serve complementary functions, with TPCNNBlock refining local structure and MambaBlock supporting the estimation of broader haze distributions.

Second, both single-path variants outperform the MLP baseline while requiring fewer parameters and MACs. The observed gains are thus more plausibly attributable to the inductive biases of the specialized blocks than to an increase in model scale.

### *E. TPCNNSpike hyperparameters*

We examine the sensitivity of TPCNNSpike to the initial values of its trainable decay factor $\tau$ and firing threshold $V_{\text{th}}$, as well as to the number of neighborhood-update iterations. Each factor varies independently around the default configuration. The training and evaluation protocol is identical to that of the component ablation.

TABLE V SENSITIVITY TO TPCNNSPIKE INITIALIZATION AND ITERATION COUNT

| $\tau$ | $V_{\text{th}}$ | Iterations | BRISQUE↓ | NIMA↑ |
|---|---|---|---|---|
| 0.3 | 0.7 | 10 | 24.5346 | 4.1456 |
| 0.5 | 0.7 | 10 | 25.0334 | 3.8973 |
| 0.7 | 0.7 | 10 | **24.2273**[a] | **4.3781** |
| 0.7 | 0.3 | 10 | 24.8741 | 3.4575 |
| 0.7 | 0.5 | 10 | 26.1589 | 3.1256 |
| 0.7 | 0.7 | 4 | 24.6268 | 4.1128 |
| 0.7 | 0.7 | 7 | 25.2371 | 3.8132 |

a. The best results are highlighted in bold.

Table V exhibits a coherent trend around the selected configuration: reducing the initial value of either $\tau$ or $V_{\text{th}}$ degrades both perceptual metrics relative to the default setting. This behavior is consistent with the TPCNNSpike update rule. A smaller $\tau$ attenuates the membrane state retained from earlier rounds and thereby limits temporal accumulation, whereas a lower $V_{\text{th}}$ facilitates firing and may allow less selective neighborhood responses to enter the coupling term. The selected pair therefore provides a useful balance between information retention and firing selectivity. Although both parameters are subsequently optimized, their initial values establish the dynamical regime from which learning begins. This indicates that initialization affects the subsequent optimization dynamics.

The iteration sweep supports the same interpretation. Ten updates outperform both shorter settings, indicating that the firing information requires sufficient rounds to propagate beyond the immediate Gaussian-weighted neighborhood. The difference between four and seven iterations is not monotonic, however, suggesting that propagation depth interacts with membrane accumulation and firing behavior rather than acting as an independent scaling factor. Overall, the results favor a regime with adequate state retention, selective firing, and a sufficiently long propagation horizon.

### F. *DegModel training source*

To determine how the source data used to learn haze formation affects adaptation, we train DegModel on three datasets while keeping the dehazing network and adaptation procedure unchanged. The resulting adapted models are evaluated on RTTS, as shown in Table VI.

TABLE VI EFFECT OF THE DEGMODEL TRAINING SOURCE ON RTTS ADAPTATION

| Datasets | BRISQUE↓ | NIMA↑ |
|---|---|---|
| O_RESIDE_6K | 29.8236 | 4.1235 |
| RESIDE_6K | 28.2335 | 4.4532 |
| NH-HAZE | **27.7184**[a] | **4.8739** |

a. The best results are highlighted in bold.

NH-HAZE yields the strongest result on both perceptual metrics. This suggests that DegModel benefits less from restricting the training source to outdoor scenes resembling the evaluation domain than from data that more clearly represent the forward haze-formation process. In particular, the physically generated, spatially nonhomogeneous haze in NH-HAZE may provide a more informative relationship between scene content and degradation, enabling the learned consistency signal to transfer to naturally hazy images.

The comparison between RESIDE_6K and its outdoor-only subset supports a related interpretation. Including the broader set of scenes improves the adapted result, indicating that diversity in the degradation-learning data may be more valuable than restricting the source to images that resemble the outdoor target domain. These findings motivate the use of NH-HAZE for the main configuration and suggest that the quality and diversity of haze-formation examples are important when training the degradation model.

## V. DISCUSSION

Our results raise several questions beyond which model obtains the best metric. Why does a local–global structure work well for haze? What does the spiking-inspired path contribute when its outputs remain full precision? Why can a degradation model trained on NH-HAZE adapt the dehazer to other real-image datasets? In this section, we discuss these questions and clarify what conclusions we draw from the current experiments.

### A. *Why do local and global features work better together?*

Haze is neither a purely local nor a purely global degradation. Edges, textures, and the boundaries between foreground and background require local processing. At the same time, deciding whether a low-contrast region is distant content or residual haze often requires information from a much larger part of the scene. A local model may preserve details but fail to recognize a broad haze veil, whereas a global model may understand the overall scene yet weaken small structures.

This explains the pattern in our ablation results. TPCNNBlock provides the larger individual improvement because local structures remain fundamental to image restoration. MambaBlock also improves the baseline while using the fewest parameters, showing that scene-wide information can be introduced economically. More importantly, the complete model outperforms both individual variants. We therefore do not view the two paths as competing feature extractors. Instead, the local path tells the network what should be preserved at a particular position, while the state-space path helps it decide how that position relates to the haze distribution of the whole scene. The learned gate then allows these two kinds of evidence to be combined throughout the network. Since the ablation compares complete configurations rather than equal-capacity modules, we use it to explain this complementarity rather than to claim that one block is universally better than another.

### B. *What does TPCNNSpike contribute?*

In our network, TPCNNSpike acts as an iterative spatial communication process. During one update, each unit receives firing information from a Gaussian-weighted neighborhood. Repeating the update allows this information to move beyond the immediate neighborhood. The firing state determines when neighboring responses affect a unit, but the feature passed to the next layer remains full precision. Thus, TPCNNSpike uses the selective and stateful behavior of spiking dynamics without forcing the image features into binary spike trains.

The hyperparameter results give some insight into this process. Ten iterations work better than four or seven, so a very short propagation is not enough in our setting. Lowering $V_{\mathrm{th}}$ also reduces performance, possibly because too many units fire and the neighborhood interaction becomes less selective. A smaller $\tau$ retains less information from earlier rounds. These trends are not monotonic, and we do not assume that more iterations must always be better. What matters is the balance between retained information, firing selectivity, and the number of propagation rounds.

### *C. Why does degradation reconstruction help real-image adaptation?*

The synthetic-trained model already produces a plausible clear image, but real photographs still contain haze patterns and color shifts that are not well covered by synthetic pairs. DegModel lets us use the real hazy input without inventing a clear target. Given a predicted clear image, we apply the degradation extracted from the input and ask whether the result can reconstruct that input. The reconstruction error then tells HazeSpikeMamba how its prediction should change.

We update only the final MambaSpikeBlock and reconstruction layer. The early features learned from synthetic pairs are kept, while the layers closest to image reconstruction are allowed to adjust to real haze. This restricted update improves all reported metrics on RTTS, URHI, and HSTS. It suggests that at least part of the domain gap can be corrected near the output stage, without relearning the whole dehazing network. The very small learning rate is suitable for the same reason: adaptation is making a limited correction to an existing solution.

NH-HAZE yields the strongest DegModel among the three training sources evaluated for RTTS adaptation. This result is useful because DegModel needs to learn how haze is formed, not only the appearance of outdoor hazy photographs. NH-HAZE contains physically generated, nonhomogeneous haze and provides a clearer relation between scene content and degradation. That relation transfers well to the target sets. We observe a related pattern when RESIDE_6K outperforms its outdoor-only subset: a more diverse degradation-learning set can be more useful than restricting the source to outdoor scenes that resemble the evaluation domain.

### *D. What do the real-image results demonstrate?*

The real-image experiments address the central question of this study: whether adaptation improves dehazing when paired clear targets are unavailable. Because RTTS, URHI, and HSTS provide no such references, PSNR and SSIM are not applicable; we instead evaluate the outputs with BRISQUE, NIMA, and SSEQ. BRISQUE and NIMA improve after adaptation on all three datasets, and SSEQ shows the same trend on RTTS. The consistent direction of these metrics provides broader evidence than any single score alone because the metrics capture different aspects of perceptual quality. The visual comparisons reinforce this pattern, showing improved visibility in heavily hazed regions without conspicuous new artifacts.

These gains should be interpreted in light of the adaptation protocol. Each unlabeled target set is used to obtain one shared checkpoint, after which its images are processed by ordinary forward inference. The method is therefore dataset-level and transductive: it neither optimizes a separate model for each image nor transfers to the target domain without seeing target images. This protocol is most suitable when a collection of images comes from a common domain, allowing the adaptation cost to be amortized across the collection.

The computational figures describe a similar trade-off. With 2.02M active parameters and 13.27G MACs, the forward network is compact relative to most deep models in the comparison. These figures measure model size and nominal computation, however, rather than hardware-level latency or energy use; TPCNNSpike still performs ten sequential updates. Adaptation adds approximately $10N$ patch updates for a target set of $N$ images, but this is a one-time dataset-level cost rather than an expense incurred for every subsequent inference.

## CONCLUSION

In this work, we have presented HazeSpikeMamba, a compact framework for single-image dehazing under synthetic-to-real domain shift. The network combines TPCNNSpike with an MHSA-free attentive state-space module in a multi-scale U-Net. TPCNNSpike propagates full-precision responses through Gaussian-weighted spatial neighborhoods over successive iterations, while the state-space path uses ASE and SGN to model global information. A learned gate fuses the two paths at every stage. The forward network contains 2.02M active parameters and requires 13.27G nominal MACs.

For adaptation to real scenes, a frozen degradation network trained on paired NH-HAZE data re-synthesizes the observed haze from the dehazed prediction. Starting from a dehaze model pretrained on paired synthetic data, the reconstruction error updates only the final restoration layers without a haze-free target during adaptation. One shared checkpoint is adapted to each complete unlabeled target set; the procedure is therefore dataset-level and transductive rather than zero-shot or per-image optimization. It improves BRISQUE and NIMA on RTTS, URHI, and HSTS, as well as SSEQ on RTTS. The ablations support the complementary contributions of TPCNNBlock and MambaBlock. Among the three DegModel training sources evaluated on RTTS, NH-HAZE gives the best adapted result.

HazeSpikeMamba therefore provides a compact approach to real-image dehazing, but the conclusions remain limited by the transductive protocol, reliance on no-reference quality metrics, and the small size of HSTS. Future work should compare TPCNNSpike with grouped-scan activations under equal model capacity, measure runtime and energy consumption on target hardware, and examine the identifiability of the degradation-consistency objective. Larger real-image datasets with reliable references will also be important for determining whether the observed perceptual gains correspond to faithful haze removal.